\documentclass[11pt,a4paper]{article}
\usepackage[utf8x]{inputenc}
\usepackage[T1]{fontenc}
\usepackage{natbib}
\usepackage{mathptmx} 
\usepackage[pdftex]{hyperref} 
\hypersetup{colorlinks,linkcolor=blue, citecolor=red}
\usepackage{calc} 
\usepackage{enumitem} 
\usepackage[a4paper, lmargin=0.16\paperwidth, rmargin=0.16\paperwidth, tmargin=0.11\paperheight, bmargin=0.11\paperheight]{geometry} 
\usepackage[all]{nowidow} 
\usepackage[protrusion=true,expansion=true]{microtype} 

\usepackage{xcolor}

\usepackage{sectsty}
\subsubsectionfont{\normalfont\itshape} 

\usepackage{titling}

\title{Post-Workshop Report on \\
Science meets Engineering in Deep Learning \\
NeurIPS 2019, Vancouver}

\author{
Levent Sagun\thanks{Facebook AI }
\and 
Caglar Gulcehre\thanks{DeepMind}
\and 
Adriana Romero\footnotemark[1]
\and
Negar Rostamzadeh\thanks{Google Brain}
\and
Stefano Sarao Mannelli\thanks{Institut de Physique Th\`eorique - CEA Saclay}
}

\thanksmarkseries{arabic} 

\begin{document}

\maketitle


\begin{abstract}

Science meets Engineering in Deep Learning took place in Vancouver as part of the Workshop section of NeurIPS 2019 \cite{sedl2019}. As organizers of the workshop, we created the following report in an attempt to isolate emerging topics and recurring themes that have been presented throughout the event. 

Deep learning can still be a complex mix of art and engineering despite its tremendous success in recent years. The workshop aimed at gathering people across the board to address seemingly contrasting challenges in the problems they are working on. As part of the call for the workshop, particular attention has been given to the interdependence of architecture, data, and optimization that gives rise to an enormous landscape of design and performance intricacies that are not well-understood. 

This year, our goal was to emphasize the following directions in our community: (i) identify obstacles in the way to better models and algorithms; (ii) identify the general trends from which we would like to build scientific and potentially theoretical understanding; and (iii) the rigorous design of scientific experiments and experimental protocols whose purpose is to resolve and pinpoint the origin of mysteries while ensuring reproducibility and robustness of conclusions. 
In the event, these topics emerged and were broadly discussed, matching our expectations and paving the way for new studies in these directions. 

While we acknowledge that the text is naturally biased as it comes through our lens, here we present an attempt to do a fair job of highlighting the outcome of the workshop.

\end{abstract}

\section{Introduction}

\subsubsection*{What do we mean by Science and Engineering} 

Every year, machine learning community shares thousands of 
articles through various conferences and journals. We expect that the work we produce and share in our circles is scientifically sound and robust. At the same time, we acknowledge that the models that are deployed based on these works have a tremendous impact on technological and social areas. Addressing this connection has never been more critical.

When we first announced the workshop, we made an effort to emphasize the need for deeper engagement across communities, in particular, the need for a platform to communicate across groups with drastically different focuses in mind. However, we deliberately omitted a definition of science and engineering much like the same way we didn't go into the applied vs theoretical distinction. Here we would like to clarify our approach: 

Each context draws its own boundary that separates seemingly opposite approaches. It may do so by specifying the goals, or by the way it utilizes existing methods. Our title is an attempt to reflect the joys of increasing the bandwidth in communication with the other side, no matter which side one begins with. The sessions are not separated according to what approach they take; rather, each session has a concrete problem(s) at hand which presents and discusses how different approaches may help tackle the problem. 

\section{Structure and Content}

The workshop consists of three themed sessions composed of invited talks and short panel discussions, followed by a panel and contributed talks covering a wide variety of methods and applications.


\subsection{Theory}

Theoretical approaches in understanding the dynamics of neural networks have started in the 90s with early results on the perceptron. In the 2010s, we have seen progress in deep linear models and past few years marked a significant development in deep non-linear models. Questions on statistical properties and training dynamics of infinite width limit (Yasaman), infinite-depth limit (Surya), and the role of the structure of data (Florent) have shown promising progress. However, in many ways, the developments left many more questions to be answered. In the theory session, we invited a group of researchers to address current challenges in this area.

The Theory Session was moderated by \textbf{Lenka Zdeborov\'a}, Institut de Physique Théorique of Saclay, and discussions have been curated together with \textbf{Adji Bousso Dieng} and \textbf{Joan Bruna}.

{\textbf{Surya Ganguli}}\footnote{\href{https://slideslive.com/38922451/an-analytic-theory-of-generalization-dynamics-and-transfer-learning-in-deep-linear-networks}{Surya Ganguli's talk \textit{An analytic theory of generalization dynamics and transfer learning in deep linear networks}}} from Stanford University presented “An analytic theory of generalization dynamics and transfer learning in deep linear networks”\cite{saxe2013exact,saxe2019mathematical,lampinen2018analytic} were he shows how remarkable phenomena observed by practitioners in deep architectures emerged already in the study of deep linear models. 

{\textbf{Yasaman Bahri}}\footnote{\href{https://slideslive.com/38922450/learning-with-realistic-synthetic-data}{Yasaman Bahri's talk \textit{Tractable limits for deep networks: an overview of the large width regime}}}, researcher at Google Brain, in “Tractable limits for deep networks: an overview of the large width regime” \citep{lee2019wide} analyzed the large width limit of a neural network. In the talk, she showed the advantages of coarse-graining over the parameters space and focusing on understanding the phenomenology in functional space.  

Finally, {\textbf{Florent Krzakala}}\footnote{\href{https://slideslive.com/38922450/learning-with-realistic-synthetic-data}{Florent Krzakala's talk \textit{Learning with "realistic" synthetic data}}}, from École Normale Supérieure, explained in “Learning with "realistic" synthetic data”  \citep{goldt2019modelling} the limitation of the most common techniques for generating synthetic datasets and proposed new models to go beyond such limits.

\subsubsection*{Discussion}
Overfitting, optimal stopping, double-descent curves, expressivity and learning dynamics in deep and wide networks are some of the many aspects that theoretician are tackling in recent years. They are different pieces of the puzzle that we must solve in order to develop a solid understanding of how deep models work.
All those phenomena were studied in detail in models complex enough to capture those aspects and simple enough to be studied and provide insights. The key takeaway is that most interesting phenomena already appear in very simple, even linear models. And a deeper understanding of those cases shed light into what happens in complex cases. 

While simple models can be enlightening in the dynamics of learning algorithms, being aware of the limitations of simple models is equally important. Where do they fail? How can we adjust it so that it doesn't? As an example in modelling, we know that linear models in large dimension are affected only by the first two moments of their data distribution, therefore they fail to capture the impact of higher orders that may be influential in complex models. On the data front, most theoretical work limits modelling of data with independent priors that doesn't lead to any particular structure, hence we need better modelling of the data that is treatable. Putting forward and understanding the limitations of the model and the assumptions adopted are fundamental steps to build a solid theory of ML and to improve the communication with other communities.

Since the field of machine learning is continually expanding at an increasing pace, the role of theory is often questioned in the community. The speakers have addressed this question and discussed four ways theory contributes to the field: it gives prescriptive principle to guide practitioners, it provides comfort with rigorous guarantees, it brings suggestions, and it brings conceptual understanding \cite{breiman1995reflections}. However, the time to develop new theoretical tools causes often a delay between discoveries of practitioners and their theoretical understanding, but the systematic application of the scientific method to carry out theoretical analysis shall help the community in bringing clarity of thought in the dust of the results.

\subsection{Vision}

The last decade has witnessed substantial progress in many computer vision tasks. This progress has been in part enabled by the introduction of large scale benchmarks, such as ImageNet, which have led researchers to explore and develop successful approaches in relatively constrained scenarios. However, the visual world is rich and complex and offers interesting challenges ahead. Moving towards real-world vision has been a recurrent topic at SEDL 2019. What does real-world vision entail? What does the path towards real-world vision look like? How can the spectrum of hypotheses involving theory, application and engineering help us navigate the upcoming challenges?

The Vision session was moderated by \textbf{Natalia Neverova}, FAIR, and discussions have been curated with \textbf{Ilija Radosavovic} and \textbf{Riza Alp Guler}.

The talk of \textbf{Carl Doersch} from DeepMind on "Self-supervised visual representation learning" exposed the importance of building representations that do not rely on huge amounts of annotated data and that are useful for downstream tasks, highlighting the potential of self-supervised learning to address problems beyond ImageNet \cite{doersch2015unsupervised,henaff2019,vandenOord2018}.

The talk of \textbf{Raquel Urtasun} from Uber ATG on "Science and Engineering for Self-driving" addressed the importance of explainability when building safety-critical systems, and argued for a holistic approach to specific applications as well as exploiting application-dependent prior knowledge, that can benefit from a theoretical explanation. \cite{Liang_2019_CVPR,Zeng_2019_CVPR,Homayounfar_2019_ICCV,bai2018}.

Finally, the presentation of \textbf{Sanja Fidler} from Nvidia AI talk emphasized the role of the data when aiming to achieve high-end performance in many specific applications, and discusses how to reduce the data annotation burden, by introducing semi-automatic approaches to obtain annotations efficiently \cite{acuna2018efficient,LingGKCF19}.

\subsubsection*{Discussion}

Researchers seek to study real-world vision utilizing challenging computer vision applications such as self-driving cars or embodied systems. Tackling such applications may require designing systems, which build representations that capture relevant semantic properties and are useful for downstream tasks, without relying on vast amounts of human-annotated data. Broadly speaking, these systems should be able to generalize to the unknown and adapt to any new, dynamic environment -- that is they should be able to act under all possible conditions and anywhere. Moreover, they should be able to reason about the 3D world and understand intent and causality. Besides the previously mentioned objectives, striving for system robustness and explainability is crucial for any safety-critical application. 

During the workshop, those points were discussed as potential steps were proposed to address some of the above mentioned challenges. On the one hand, the human-annotation burden could be mitigated by making simulators more realistic, building semi-automatic annotation tools, or exploiting self-supervision and auxiliary data modalities as supervision signals. On the other hand, representations could be improved by building systems that leverage general knowledge about the world (e.g. laws of physics) or application-specific priors, which may help uncover relevant but otherwise invisible variables.

Presenters proposed potential future directions that would help enhance the synergy between theory, applications and engineering. First, what is the meaning of big data in the context of real world vision? Researchers hypothesize that understanding (i) why systems work on current datasets (ImageNet), (ii) how the amount of data and annotations affects the performance of the system, and (iii) how to perform data collection more efficiently (e.g. by means of very realistic simulated environments or semi-automatic data annotation tools), might shed some light. Second, how could we solve the problems with less data? Researchers hypothesize that (i) designing suitable loss functions and regularizers, (ii) exploiting good data pre-processing techniques, and (iii) including prior knowledge, could lead to improving the data efficiency of the systems. Last, how could we better assess the robustness of the designed systems? In this case, researchers hypothesize that defining application-specific safety bounds and improving the understanding of adversarial examples could be two potential research avenues.

\subsection{Further Applications}
Deep learning has made an early impact on application areas such as computer vision \citep{krizhevsky2012imagenet} and speech recognition \citep{hinton2012deep}, however, applications on other domains such as NLP, robotics, and self-driving cars followed later on. Such widening of application areas created unique tasks to tackle. The data distributions, the structure of the problem, the objective functions and modalities of the problem presents exciting scientific and engineering challenges. Some of these challenges have a workaround developed by practitioners but scientifically not very well-studied such as the dataset generation and the state of art models like BERT \citep{devlin2018bert}. Some interesting questions like "How should be a proper benchmark for an NLP task generated?", "When the theory can be helpful for the applications?", and "How do we detect overfitting in generative models?" were raised during the talks.

{\textbf{Yann Dauphin}} from Google Brain moderated the further applications session and the discussions have been curated with the support of \textbf{Orhan Firat} and \textbf{Dilan Gorur}.

The talk {\textbf{Douwe Kiela}}\footnote{\href{https://slideslive.com/38922445/benchmarking-progress-in-ai-a-new-benchmark-for-natural-language-understanding}{Douwe Kiela's talk \textit{Benchmarking Progress in AI: A New Benchmark for Natural Language Understanding}}} from FAIR has focused on the adversarial NLI and how to create new natural language benchmarks that require more robust models \citep{nie2019adversarial}. During the talk Douwe discussed the shortcomings of typical machine learning benchmarks, and how they can be exploited easily. Instead, dynamic benchmarks that can be extended by adopting an adversarial strategy. 

{\textbf{Audrey Durand}}\footnote{\href{https://slideslive.com/38922458/trading-off-theory-and-practice-a-bandit-perspective}{Andray Durand's talk \textit{Trading off theory and practice: A bandit perspective}}}, from Université Laval and MILA, discussed the trade-off of theory and application in bandits \cite{durand18a,lupu2019leveraging}. UCB was favored for their theoretical guarantees, however, some practitioners used Thompson sampling, a method with sparse theoretical results until 2011. Further examples have been discussed on how such developments made a big impact on the applications of the method and the other way around as well.

{\textbf{Kamalika Chaudhuri}}\footnote{\href{https://slideslive.com/38922459/a-three-sample-test-to-detect-data-copying-in-generative-models}{Kamalika Chaudhuri's talk \textit{A Three Sample Test to Detect Data Copying in Generative Models}}} from University of California San Diego presented a recent work on a non-Parametric test to detect data-copying and overfitting in generative models \citep{meehan2020non}. Kamalika first discussed how overfitting happens in generative models and proposed a method to detect overfitting in generative models by using 3-sample test instead of using methods similar to Mann-Whitney test.

\subsubsection*{Discussion}

Although the limits of applied deep learning techniques are unclear, those methods are (undeniably) widely used and full surprises. Such potentially powerful tools consequently create a competitive environment across domains. And each area has its own set of theoretical and engineering challenges. Common (and recurring) themes of challenges across disciplines emerge: evaluation of proposed models relying on test data, the concept of under/overfitting for a given model, and the reliability-popularity trade-off of models with weak theoretical guarantees.   

Typically, problems that move beyond their infancy have their fixed benchmark datasets for which new models are evaluated. Even though benchmark datasets can be renewed, the number of proposed models far exceeds the number of existing benchmarks. This abundance of models in comparison to existing datasets leads to a competitive publication environment where breaking records on a given task is the sole goal. On the other hand, for every given model, one can adversarially modify the data in such a way that the update can break the model. Such fragility may imply either that the models are not good enough, or alternatively, the framework in which we test our models is not good enough. If it is the latter, we can change the way we curate data more frequently and adopt a notion of dynamic datasets along with reproducible testing pipelines. A further emphasis of focus on datasets themselves seems necessary. This change of focus will also imply changes in the way we test models and evaluate under/over-fitting.

Various areas of applications showcase the trade-off of theory and application in topics where simple, yet useful, abstractions have wide-ranging engineering applications. On the one hand, part of the community may wait for convergence guarantees before widely deploying certain models even though they are demonstrated to work in practice. On the other hand, the existence of working models also exemplifies how prior empirical findings help shape theoretical efforts, thereby once again emphasizing the value of communication across groups.

\subsection{Panel - Interaction between science and engineering}

The massive expansion of computer science as a field gave rise to a considerable number of sub-communities that developed their own vocabulary and their methods. Nowadays, the separation of those sub-communities, and the consequent communication issues, are at the basis of several problems that affect the whole community. This can be seen in: old results often being independently rediscovered; and in the struggle of peer-reviewing on topics at the cross-section of two or more sub-communities. All these problems are aggravated by the unsustainable demand for publications in the field, that segregate the communities even further and lower the overall quality of the publications. 

During the panel, the lack of communication across these sub areas has been discussed, and it was pointed out that there is a need for a joint effort in reducing such distances across sub-communities. We should make an effort to increase awareness for the works that appear in our sub-disciplines while maintaining the identity and purpose of each group. An effort is needed in reducing the number and improving the quality of publications, as well. Everyday, it becomes harder to keep track of the publications, and the fast reviewing process is pauperizing the quality of the publications. This leads to a situation in which it gets harder to trust published results which also results in reproducibility issues. An idea that will help in this direction is to adopt open-reviewing strategies. Eventually adding a layer on-top of arXiv by giving the possibility to comment and reply in preprints might alleviate some of the troubles.

{In the discussion} 
\textbf{Finale Doshi-Velez} (Harvard), \textbf{Surya Ganguli} (Stanford), \textbf{Been Kim} (Google Brain), \textbf{Aparna Lakshmiratan} (Facebook), \textbf{Jason Yosinski} (Uber) intervene, and the panel was moderated by \textbf{Zack Lipton} (Carnegie Mellon University) with advisory support by \textbf{Michela Paganini}. 

\subsection{Contributed sessions}

We received 45 diverse and high-quality submissions. With the help of 34 reviewers, we evaluated the quality of the papers and managed to check their fitness to the theme of the workshop discussed in our call for papers. In the end, we selected 27 submissions for a poster presentation, among which we further selected 5 where the authors delivered an oral presentation of their work.

Call for papers invited specific types of works that either lie at the intersection of areas, furthermore we also added three more additional optional criteria: (1) negative results, (2) simplified versions of complex models from practitioners to be presented to theoretical scientists, (3) possible broader implications theoretical work addressed in a way that is accessible to the practically minded. We added these options because it can be particularly challenging to get this kind of work accepted in today's publication regime. But it is precisely this kind of work that, in our opinion, fosters the communication across said groups.
\\ \\
Here, we provide brief descriptions of 5 contributed presentations:

\textbf{Sho Yaida} from Facebook AI Research presented \textit{Non-Gaussian Processes and Neural Networks at Finite Widths} analyzing in perturbation theory the effect on finite widths on an infinitely-wide net. The study allows to go through the network flow layer by layer integrating out random variables, analogous to the renormalization-group flow.

\textbf{Jonathan Frankle} from MIT presented \textit{Training Batchnorm and Only Batchnorm} a work in collaboration with David J Schwab (ITS, CUNY Graduate Center) and Ari S Morcos (Facebook AI Research (FAIR)). In the work, the authors show a network could achieve high accuraccy in real datasets having fixed random layers and only tuning the parameters of batchnorm.

\textbf{Guy Gur-Ari} (Google) presented \textit{Asymptotics of Wide Networks from Feynman Diagrams}, a joint work with Ethan Dyer (Google). In the work, the authors used Faynman diagrams to analyse the training dynamics of wide networks going beyond the large width limit by obtaining closed-form expressions for the higher-order terms.

\textbf{YiDing Jiang} from Google presented \textit{Fantastic Generalization Measures and Where to Find Them}, in collaboration with Behnam Neyshabur (Google), Dilip Krishnan (Google), Hossein Mobahi (Google Research) and Samy Bengio (Google Research, Brain Team). What is the right measure complexity measure for deep neaural network? In their work, numerous well-designed experiments are compared through various complexity measures.

\textbf{Martin Ma} and \textbf{Muqiao Yang} from Carnegie Mellon University presented their work \textit{Complex Transformer: A Framework for Modeling Complex-Valued Sequence} in collaboration with Dongyu Li (Carnegie Mellon University), Yao-Hung Tsai (Carnegie Mellon University) and Ruslan Salakhutdinov (Carnegie Mellon University). The work exploits phenomenon that are better captured through the use Fourier Tranform. In their study, the authors develop a transformer that is able to handle complex representations, achieving state-of-the-art performance.


\section{Recurring topics of debate}

\subsubsection*{Discussion on two kinds of theoretical contributions}
Throughout the sessions and especially in the panel discussion, participants emphasized differences in theoretical approaches to studying inner workings of deep models. Two overarching lines of research emerged: (1) theoretical guarantees focusing on rigorous assumptions leading to rigorous theorems, (2) complex systems approach of probing the models to test their limits and find out about their properties. A theme that emerged throughout the workshop is the limits of applicability of theory when one insists on the rigor to an extent that it allows provability but shaves off of their capacity to explain observed phenomena. Even though participants pointed out the value of mutual existence of both approaches, the cultural clash that systematically devalues the latter approach has been criticized as it hinders progress in understanding the phenomenon.

\subsubsection*{Broader impact of theoretical studies}

Impact of theory has been discussed in two core areas: (1) limitations of theoretical results, (2) effect of theory in practical applications. In some sense, both questions are related to a notion of the boundary of applicability. In the former case, once the necessary restrictive assumptions have been lifted, things may fall apart. In the latter, the phenomenon may hold for a broader set of not-yet-proven cases. A common theme across the sessions has been about the idea of researchers being more aware and thoughtful of the boundaries of their work.

\subsubsection*{The importance of toy models}
Science/theory builds upon simple models. They are on the one hand useful for analytical studies, but on the other hand, they help to identify relevant factors contributing to phenomena of interest. A model can also come from limiting situations of real architectures, for instance infinitely wide or infinitely deep networks may be far from reality but can already capture part of the phenomenology. In this limit, we can understand more and eventually move back to the practical cases by expanding around the limits. An example of this is the current strong interest around the neural tangent kernel limit. It may be true that this limit does not capture the reality, but the large number of results collected in that set-up may pave the fundamental to the next discovery. 

\subsubsection*{Understanding structure in data}
Data, architecture and algorithm are the three pillars for funding the theory of machine learning. Their study and, in particular, the study of their interactions is a necessity to step forward. So far, a huge effort has been put to study architectures and algorithms, now we need to study data better. All these directions are crucial, but the role of data is the least explored. Many applications rely on a large amount of data that in many cases, we can hardly rely on automatic labelling. A better understanding will have a dramatic impact on the whole community. Better insights, for instance, allow building more realistic simulators or designing better priors to reduce the number of data in training.

\section{Conclusions}

In the last decade, the mathematical theory behind ML did not match the pace of the progress in its applications. We argue that the use of scientifically sound but non-rigorous methods will bring an important contribution to reduce the distance between those communities, in particular the role of physics. In its history, physics developed a \textit{forma mentis} that helped in dealing with complex problems. The mindset and several (non-rigorous) tools of physics can be applied in ML as they have been applied to other (natural) phenomena. In particular, physics/physicists help the community by placing themselves in between mathematicians/theoretical computer scientists and engineers, in the abstract line that connects pure science to empirical sciences or engineering. As a community, we should sustain the connection between theory and applications. Fundamental knowledge gives prescriptive principle to guide practitioners, brings suggestions and conceptual insides. All applied branches of ML would benefit from those points, in particular, critical applications where good performances of ML are not enough for its adoption but where explainability is needed, for instance, for accountability.

The complexity of the visual world, textual world, spaces of exploration and as such are widely complex, which hinders theoretical progress in the respective fields. Yet, many communities have successfully explored approaches that work well on various previously challenging tasks. The goal we seek to reach is to put such advances to a solid grounding, thereby help develop further progress.

SEDL 2019 workshop gathered different perspectives, from researchers and practitioners, to address seemingly contrasting challenges. The role of science and engineering was largely discussed in all the sessions, concluding that science and engineering are intertwined, and their symbiotic interaction is beneficial to push the boundaries of research in both theory-driven and application-driven sub-fields. However, to truly benefit from the diversity of thought that arises across such sub-fields, it is important to establish effective communication strategies that mitigate the existing language barriers. Thinking of a continuous spectrum of interests across sub-fields could help us improve communication and hopefully lead to more impactful contributions.

\section*{Acknowledgements}
The authors of this text assume all the responsibility for the opinions selected and presented in the present report. However, we would like to take the chance to thank everyone involved in the making of the workshop. We thank Yasaman Bahri whose contributions shaped the beginning of the workshop. Our advisors, Joan Bruna, Adji Bousso Dieng, Ilija Radosavovich, Riza Alp Guler, Orhan Firat, Dilan Gorur, and Michela Paganini, helped curate the talks and discussion points for each session. Our moderators, Lenka Zdeborova, Natalia Neverova, Yann Dauphin, and Zack Lipton, helped bridge the talks and discussions. Our speakers, Surya Ganguli, Yasaman Bahri, Florent Krzakala, Carl Doersch, Raquel Urtasun, Sanja Fidler, Douwe Kiela, Audrey Durand, Kamalika Chaudhuri, delivered excellent seminars. And finally, our panelists, Finale Doshi-Velez, Surya Ganguli, Been Kim, Aparna Lakshmiratan, Jason Yosinski contributed intriguing discussions. Our speakers for the contributed talks, Sho Yaida, Jonathan Frankle, Guy Gur-Ari, YiDing Jiang, Martin Ma, Muqiao Yang.
We are glad to hear that all the discussions hit right at the heart of the theme of the workshop.

\bibliographystyle{plain}
\bibliography{report}

\end{document}